\documentclass{article} 
\usepackage{iclr2016_conference,times}
\usepackage{hyperref}
\usepackage{url}
\usepackage{epsfig}
\usepackage{graphicx}
\usepackage{amsmath}
\usepackage{amssymb}
\usepackage{algorithmic}
\usepackage{algorithm}
\usepackage{memhfixc}
\usepackage{amstext}
\usepackage{stmaryrd}
\usepackage{fancyhdr}
\usepackage{booktabs}
\usepackage{url}

\title{Formatting Instructions for ICLR 2016 / \\ Conference Submissions}

\author{Anirudh Goyal\\
Center for Visual Information Technology\\
International Institute of Information and Technolgy, Hyderabad\\
\texttt{anirudh.goyal@students.iiit.ac.in} \\
\And
Marius Leordeanu \\
Institute of Mathematics of the Romanian Academy \\
21 Calea Grivitei, Bucharest, Romania \\
\texttt{leordeanu@gmail.com}
}

\begin{document}

\title{Stories in the Eye: Contextual Visual Interactions
for Efficient Video to Language Translation}
\maketitle

\begin{abstract}

Integrating higher level visual and linguistic interpretations is at
the heart of human intelligence.
As automatic visual category recognition in images is approaching human performance,
the high level understanding in the dynamic spatiotemporal domain of videos and its translation
into natural language is still far from being solved. While most works on vision-to-text translations
use pre-learned or pre-established computational linguistic models, in this paper we present
an approach that uses vision alone to efficiently
learn how to translate into language the video content. We discover, in simple form, 
the story played by main
actors, while using only visual cues for representing objects and their interactions.
Our method learns in a hierarchical manner higher level representations for recognizing subjects, actions
and objects involved, their relevant contextual background and their interaction to one another over time.
We have a three stage approach:
first we take in consideration features of the individual entities at the local level of appearance, then
we consider the relationship between these objects and actions and their video background, and third,
we consider their spatiotemporal relations as inputs to classifiers at the highest level of interpretation.
Thus, our approach finds a coherent linguistic description of videos
in the form of a subject, verb and object based on their role played in the overall visual story
learned directly from training data,
without using a known language model.
We test the efficiency of our approach on a large scale dataset containing YouTube
clips taken \emph{in the wild} and demonstrate state-of-the-art performance, often
superior to current approaches that use more complex, pre-learned linguistic knowledge.
\end{abstract}

\section{Introduction}

The connection between visual interpretation and linguistic expression is at the heart of our mind - it is a link incompletely
understood in science and artificial intelligence today. The association between vision and language bridges our perception of the world to the way we communicate with each other. Solving this
scientific challenge from a computational perspective would have a significant impact in the way we understand our own way of thinking. 
Moreover, it would help in the development of a wide variety of new technologies with great potential to improve the quality of life.
We, humans, are able to effortlessly interpret a visual scene or event and then describe it concisely in just a few words. The usual case is that of a subject
performing an activity and interacting with objects or other subjects.
From a quick glance at the visual world, we immediately understand the essence of what is going on around us: from seeing 
objects, interpreting their interaction
and the overall setting, or scene of the story, to predicting activities, intentions and the future course of events.  Often we agree on the high level description of events that occur in a given video sequence, despite the natural variation in our individual experiences and points of view.
When we do not use exactly the same words to describe a situation, we usually use
very related ones that describe semantically similar events. This is exactly the problem we tackle here.
Can we have an automatic way to learn directly from the raw visual data to classify different subjects, objects and verbs
that describe short events?  Can we take in consideration their interactions and relation to the contextual scene in a pure
visual sense, without using a language model, in order to provide an accurate linguistic description in the form of Subject-Verb-Object triplet?
Learning to translate visual data into basic (subject, verb, object) translations
is an interesting challenge, a first step that
could shed some light on the possibility of learning natural language directly from vision.

The task of describing images or videos with text has received a growing attention during the last few years.
Most existing research that integrates vision and language has been studying the description of static
images (\cite{kulkarni2011baby,li2011composing,farhadi2010every,yao2010i2t,feng2013automatic,ordonez2011im2text}).
The problem is still far from being solved, with linguistic descriptions
usually being limited to simple and rather strict descriptions. They usually
have a single atomic action, one main actor acting upon a single object. This is also the case
we study here, our method being focused on video to (subject(S), verb(V), object(O)) translations -
a task tackled by a considerable number of recent works (\cite{thomason2014integrating,guadarrama2013youtube2text,xu2015jointly,venugopalan2014translating,barbu2012video,das2013thousand,rohrbach2013translating,senina2014coherent,yu2013grounded}).

We face three main challenges: 1) first, the problem of video classification has not been solved yet. The recognition of object classes occurring in a video, the understanding of what they do (actions) and how they relate and interact with each other, are problems that are difficult to model and compute efficiently. The vast sources of variation in illumination, changes in pose, occlusions, intra- and inter-class differences, make even object detection very hard. However, recent advances with hierarchical recognition models, such as the highly successful Deformable Part Model (\cite{felzenszwalb_ObjDetect_pami2010})
and the deep learning approaches with neural networks (\cite{DeepNet_ImageNet_nips2012,simonyan2014very,szegedy2014going})
show great promise towards solving the object categorization and detection part, fast approaching the human performance.
2) While image recognition is enjoying a great success, recognition in video is different and it is not clear yet how the spatiotemporal consistency
that is naturally present in a video could be best used for improved recognition. Moreover, the larger data available in a video contains important and relevant information about the main objects of interest as well as about the contextual scene in which the events take place. How could we best take advantage of the contextual relationships between objects, and between objects and the scene for superior classification. In video, the foreground and background can be better separated automatically (\cite{stretcumultiple}), and considering both their contrasting properties and their interdependence could benefit recognition.
3) Once we have a reasonably accurate way of recognizing subjects, objects and their potential interactions separately, it is still not clear, just from visual information alone, how we could
produce a coherent and meaningful SVO triplet
without prior linguistic knowledge.
Language inherently contains semantic relations that reflect dependencies which exist in the real world. An interesting question arises here: to what extent it is possible, just from visual information to learn these dependencies and bypass a strong linguistic model learned or known a priori, in order to recover the correct SVO triplet?

\section{Relation to prior work}

The approaches that are most similar to our method
are the recent works of (\cite{guadarrama2013youtube2text,xu2015jointly,venugopalan2014translating,thomason2014integrating}),
which are able to translate videos into SVO triplets with a relatively numerous vocabulary of concepts
learned from a large set of training videos taken \emph{in the wild}.
\cite{guadarrama2013youtube2text} propose a method based on semantic hierarchies. They take a language driven approach in which semantic relationships are learned between different labels by mining large corpus of text
and then clustering semantically related concepts based on their correlations and co-occurrences in human annotated text descriptions (\cite{pereira1993distributional}).
~\cite{thomason2014integrating} use a factor graph model that
combines state-of-the-art visual recognition systems with probabilistic knowledge mined from
large text corpora that are independent of the video dataset.
Following a similar direction,~\cite{venugopalan2014translating}
employ a two layer LSTM model on top of visual features learned with the Convolutional
Neural Network model available in Caffe (\cite{jia2014caffe}), a minor variant of AlexNet (\cite{DeepNet_ImageNet_nips2012}) 
and trained
on ILSVRC-2012 classification subset of ImageNet (\cite{russakovsky2014imagenet}).
The Long Short Term Memory model
uses the CNN features as inputs and it is, for best performance, pre-trained on other 
two large image datasets Flickr30k (\cite{young2014image}) and Microsoft COCO 2014 (\cite{lin2014microsoft}),
then fine-tuned on the video dataset used in experiments.
One limitation of this approach (\cite{venugopalan2014translating}) is that its video representation exploits only weakly temporal information, as it accumulates CNN feature responses over time by mean pooling.
While the previously mentioned works explore language modeling on a fixed visual model,
the work of \cite{xu2015jointly} goes a step further by combining a compositional semantics language model with a deep video model
into a single unified framework.

Compared to the three previous works above, we rely only on visual classifiers and model explicitly contextual relationships at several levels, 
that of individual subjects, verbs and objects, their relation to the  
surrounding spatiotemporal context (e.g. global features from the video that contains them), as well as relations 
between the different entities.
It is important to emphasise that we do not learn language statistics
from external sources and we do not model language knowledge explicitly.
All the linguistic relationships are indirectly
captured by the visual information and the contextual interaction between the video scene and the visual classifiers, which 
fire separately or together, in a video
at different relative moments in time. The visual cues we start from are
pre-trained CNN
features from the superior, second to last fully connected (fc7) layer of (\cite{jia2014caffe}).
Instead of pooling these features over the entire video (a common approach), we
also perform key frames and feature selection for modeling foreground, and thus 
capture an effective foreground-background contextual relationship.

In different work, ~\cite{yao2015describing} addresses the limitations of ~\cite{venugopalan2014translating} by employing a 3-D convnet model (which is pre-trained on different video datatsets of action recognition) in order to incorporate spatiotemporal motion features. For capturing changes and movement, they
extract dense trajectory features over the video. They also include an attention mechanism that learns to weight the 
features non-uniformly over frames, rather than blandly averaging them. We propose a simpler approach for exploiting temporal consistency in a video. Instead of using a complex LSTM model, we take advantage of co-firing of different visual classifiers, at different relative temporal sections in the video. Different from ~\cite{yao2015describing} we do not use features that ar specific for motion, such as cues computed from optical flow. All our features are trained on a per static image basis, such that accurate verb predictions are the result of the visual context from the scene and from the other classes present. Our relatively simpler model, proves its effectiveness and generally outperforms our competitors by an important margin
on a large video dataset (\cite{chen2011collecting}).

\section{Overview of our approach}

We address the porblem of linguistic video interpretation by first creating classifiers to predict
the different concepts, subject, object and verb from pure visual information that we extract from
deep AlexNet-like CNNs pre-trained on the large ILRSVC-2012 image classification dataset
(\cite{jia2014caffe,DeepNet_ImageNet_nips2012}). In particular, we use features from the last fully-connected layer (fc7).
As shown in our experiments these features
are often able to predict video content, such as the presence of specific subjects, objects
and even verbs,
without using any explicit motion information. The correct prediction of verbs is due to the fact that
higher level concepts, such as words in a sentence,
are strongly interconnected and the presence of one will
influence the expectation of another.
For example the appearance of a boy could raise the expectation for the
verb "play" and the object "ball", among other probable choices. We show that at this first level of interpretation it is important
to take in consideration mostly the key frames and only a few of the CNN features, which are more likely to belong to the foreground
than to the scene elements in the background.

\begin{figure}[t!]
\begin{center}
\includegraphics[scale = 0.3, angle = 0, viewport = 0 0 1500 690,clip]{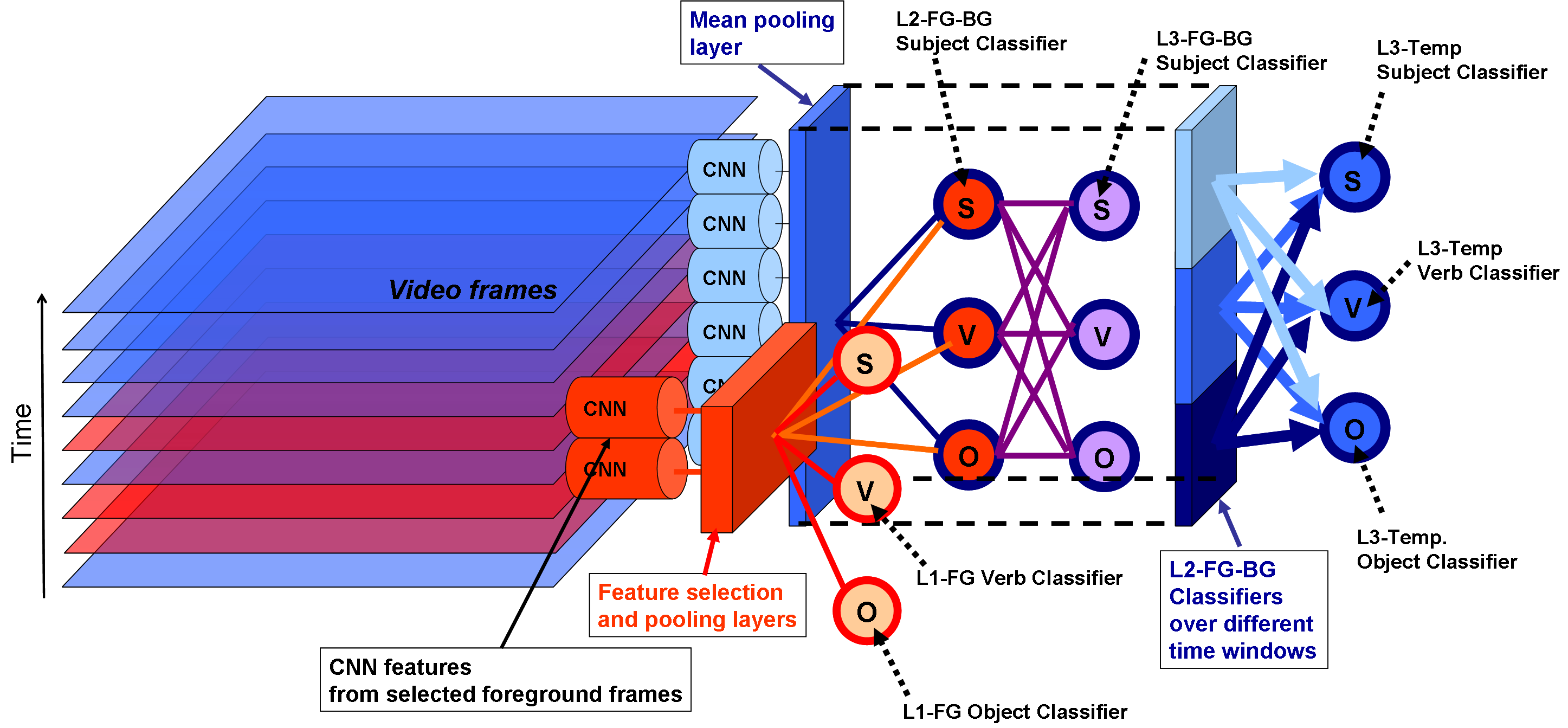}
\caption{Our \emph{visual story} system, as a model in which individual subjects, objects and the video scene interact visually to form more coherent and meaningful translations. Our system has three levels, from the detection of single entities (level L1), to capturing the relationship with the overall scene (level L2) and the interactions between the different (S,O,V) elements (level L3).
Each node is a linear classifer, trained with linear SVM, getting its inputs from the features or classifiers below in the hierarchy.
}
\label{fig:video_story}
\end{center}
\end{figure}

For the automatic discovery of frames and features that are likely to belong to the foreground objects and actions,
we propose an efficient method for unsupervised selection.
The relevant foreground frames are usually different from the average frames in the video in the sense that they have strong feature responses, and those responses also come from salient features. These salient features are also those with strong average response over time.
The discovery of salient foreground cues is
related to many works in video and image understanding, based on similar principles. For detecting foreground regions in images related ideas can be found in the works of (\cite{borji2012salient,cheng2015global,hou2007saliency}), whereas in video the supervised selection of key frames 
is known to be very effective for action and activity classification 
(\cite{ellis_rahul_ijcv2012,lv_cvpr2007,carlsson_2001,zanfir2013moving}). 
The method proposed here is unsupervised and more related to the VideoPCA approach for discovery of foreground regions 
in video (\cite{stretcumultiple}).
Thus, by leveraging information from the most representative frames and features we construct our first level of classifiers. These level 1 (L1-FG) classifiers
are then augmented with information from the background, in order 
to obtain our background enhanced L2-FG-BG second-level classifiers. 
An overview of our system is presented in Figure \ref{fig:video_story}.

Interestingly enough, even though we first need to remove the background in order to emphasize aspects of the foreground
, at a second stage,
we need to add overall scene information back. The importance of global scene context was recognized in computer vision for a long time (\cite{Oliva01}), while recent findings in psychology suggest that contextual scene information is crucial for recognition, independent of the attentional focus \cite{munneke2013influence}. This suggests that scene recognition happens before the detection of individual objects.

What is important at this stage,
is our ability to separate the foreground from the background, but both types of information
matter, in different ways, as our experiments clearly show. Thus, by augmenting our foreground descriptors with descriptors from the background we consistently and significantly improve recognition (Tables \ref{tab:most_and_any}, \ref{tab:most_and_any_WUP}). These foreground-background descriptors are the ones fed into the L2-FG-BG classifiers. The background, taken in this case as information over all features from the remaining frames, constitutes the other part of the story, the scene and context in which the subject, verb and object co-exist. While putting the whole background 
into a single pot, without separating yet information from other individual classes, we capture the scene gist - which is able to add important ingredients for the prediction of a particular subject, verb or object.

Once the foreground-background separation helped with improving recognition, we move to an even higher level of 
interpretation-in-context, in which relationships between different classes of subjects, verbs and objects can predict, at the next level of abstraction,
the presence of the same subjects, verbs and objects. 
At this level we consider the outputs of the L2-FG-BG classifiers as inputs, for the same classes,
to form level 3 classifiers, referred to as L3-FG-BG.
In particular, by taking the top $c$ class responses at level 2 for all three types, subjects, verbs and objects, we then form level 3 descriptors from these responses and learn linear
models, using linear SVM, that could predict, at level 3, from level 2 inputs. 
We show that by doing so we boost the performance, as we now take in consideration
not only the individual visual appearance of these words, but also interactions between them.
These relationships, as expected, improve the logic of the output triplets.

\begin{figure}[t!]
\begin{center}
\includegraphics[scale = 0.5, angle = 0, viewport = 0 0 690 800, clip]{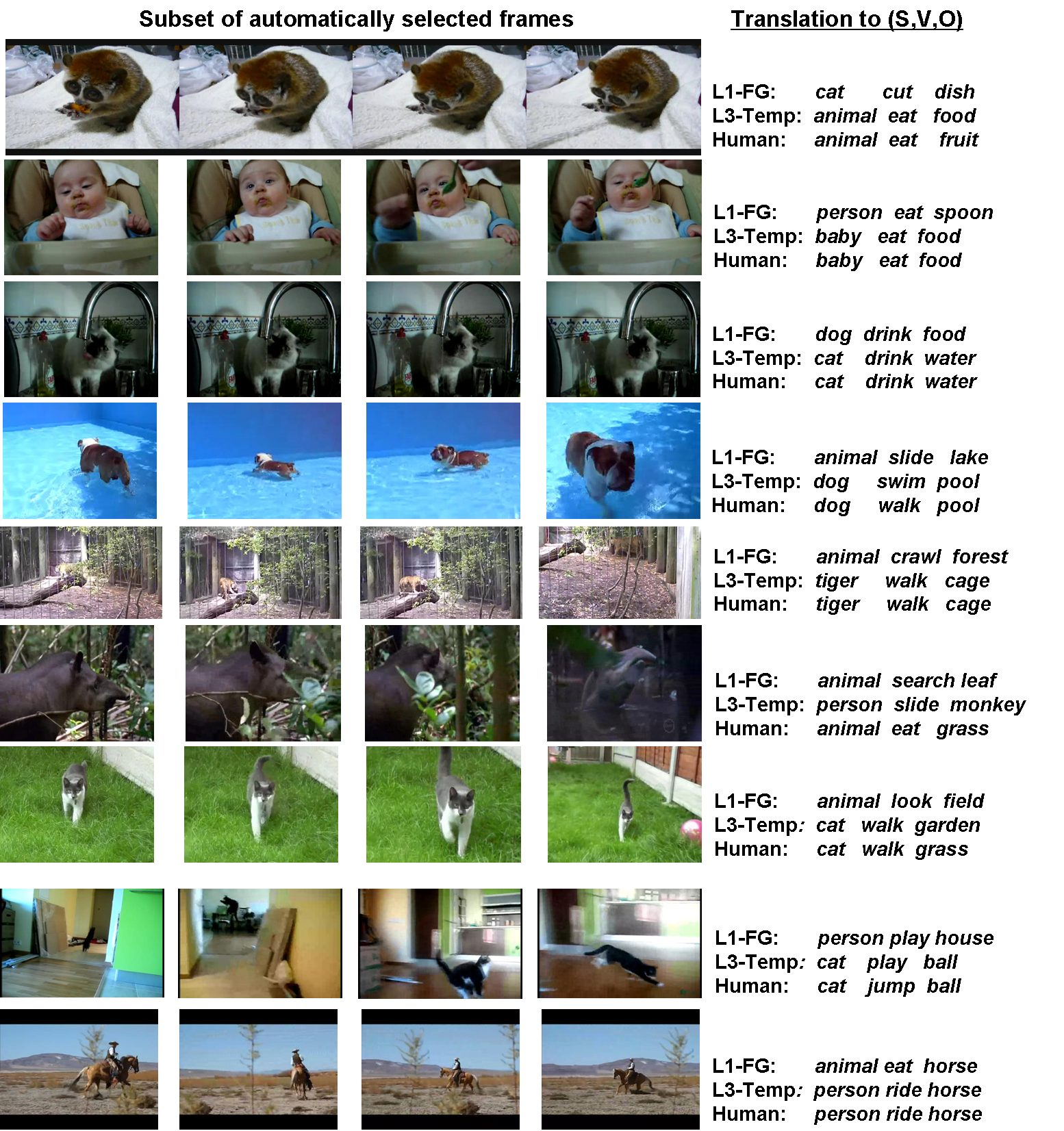}
\caption{Representative qualitative SVO translations using classifiers at different levels, L1-FG and the final, best performing L3-Temp. Note that at the third level classifications are much closer to the ground truth, as the contextual relationships between classes helps in creating a more coherent SVO triplet. }
\label{fig:L1_vs_L2}
\end{center}
\end{figure}

Let us take an example: imagine we are at the beach
with people enjoying their vacation in the sun. 
We can take a \emph{person} detector to get a few candidate locations for people in the scene - these are the L1-FG classifier responses.
Then the background, the overall scene,  could improve the recognition of people: we expect that at the beach there are many people during the summer, so, in this case, the context will encourage
the detection of people around the region belonging to the sandy beach - and these are the L2-FG-BG detections.
After we get context-improved \emph{people}, who happen to play voleyball, 
by further using L2-FG-BG classifiers of other classes, 
for verbs such as \emph{play} and
for objects such as \emph{ball}, as inputs to the next level, we could get improved L3-FG-BG
\emph{people} detections. 
This happens because now the story contains more elements,
in which people are strongly anchored into a specific \emph{role} - as context to the scene and other classifiers - and are harder to miss.
Indeed, it is harder to mistake a person for something else, when so many elements start fitting into place. Now,
at this level, the beach, the people, the ball and the game of volleyball are all part of a coherent story, which becomes a strong contextual support
for all these concepts. If we then move even higher in the contextual hierarchy and consider temporal dependencies, such as which subject or object happened first, or whether they appeared at the same time in the scene (the most common case), the recognition is further improved.
In Figure \ref{fig:L1_vs_L2} we show some representative qualitative examples of translations at different levels: the first L1-FG level and the last L3-Temporal level, which considers contextual inputs from the scene and the other objects over specific time periods. We will discuss this case 
in more detail in Sec. \ref{sec:algorithm}.

\section{Principles and Contributions of the Visual Story Approach}

Before we go into the implementation details of our method, we first lay down the basic principles and contributions of our overall approach - an attempt to understand
objects, scenes and their interactions in the context of a unified visual story. It is through this story (~\cite{schank1995knowledge,connelly1990stories,pahl2010artifactual})
that we aim to understand objects, such as their role will convincingly indicate the identity of each class, through rich contextual connections.

\textbf{The object-scene dualism:} The inter-dependence between an object and its scene is
a central idea in our approach, and we treat both the object and the scene together for recognition. Here, by \emph{object}, we mean any physical object or simple, atomic action, which could be denoted by our subjects, verbs or objects. By \emph{scene} we mean the overall dynamic and static setting in the video, the global spatiotemporal context in which actions and activities take place. 
We propose two complementary views of an entity: 1) an object, subject or verb is seen and recognized based on its own, intrinsic features such as appearance, shape, movement, or other local feature responses.
and 2) that object or verb should also be interpreted from the perspective of the overall scene. The overall larger context around that entity will also tell what that entity is. For example: the car tells that at a specific location should be a wheel. Then, the wheel-only classifier confirms (or not) that hypothesis, but only when it has sufficient information. When it does not, due to bad lighting conditions or other noises, the scene will provide the safety belt of a plausible answer to give a coherent view of the overall image. In the same way, at the higher level of an activity, the game of volleyball and the beach constitute strong indication for the presence of \emph{people}, the verb \emph{play} and the object \emph{ball}, among others.

\textbf{Classes are context to one another:} there are many examples of objects providing contextual information to other objects, for improved visual recognition. A driver could provide context for the car, if, for example, only the driver’s face and parts of the windscreen appear in the image. The car can also provide contextual input for the person, when, for instance, the driver can barely be seen when driving, due to very low resolution or bad lighting. E.g., a man is riding a horse, very far away in the field. Then, the person and horse detectors will give contextual inputs to each other and the verb \emph{riding}, and raise the
the confidence of our seeing of a man riding a horse. Thus, at the third level in our system, the first few top detections of subjects, verbs and objects based on the level 2 classifiers, become inputs to all of the potential subjects, verbs, and objects, at level 3. Note that this induces different classifiers per class, at different levels - which is an important contribution of our work, different from current approaches.

\textbf{Objects and actions play roles in a story} Each level in our contextual hierarchy, gathers information from conceptually larger basins of interpretation, to better identify the role played by each \emph{player} in the larger story. Of course that our system, presented here is a relatively simple implementation of the principle discussed here, but it is a useful and effective step in our endeavor to treat objects and actions as elements with a clear contribution to a large context, in both space, time and meaning.

\textbf{Feature selection and re-using prior knowledge is important:} the problem of efficient feature selection becomes more important as we move towards higher levels of interpretation, such as activities in video. If at the lower levels dense systems could be learned by considering all information at once, we believe that at the higher level of objects and actions there is more freedom of movement and representation, and effective feature selection could considerably reduce the exponentially growing number of possibilities. In our case feature selection is performed at two levels: first, we perform unsupervised feature selection by \emph{pooling} the salient frames and features in order to detect the potential foreground elements and cues. The same \emph{salient pooling} is performed at the next level, when we take the top detections from level 2 classifiers as inputs to the classifiers from level 3. Selecting in an unsupervised manner potentially interesting regions and features is also related to approaches in object recognition, in which segmentation cues are used in order to guide the process of attention (\cite{girshick2014rich}).

\section{Algorithm Implementation}
\label{sec:algorithm}

Our method has three stages. First, we perform feature selection and create L1-FG classifiers. At the second stage we create L2-FG-BG classifiers by considering also background information as discussed before. Third, we consider L2-BG-FG responses over the whole video, to create L3-FG-BG classifiers (by using the top $c$ detections from the previous stage L2 as inputs to stage L3). 
Another option, which considers, additionally, temporal relationships is: 
we get the level 2 classifier responses over different parts of the video (first, middle and last), get top responding classifiers for each video part and each type, subject, verb or object, as inputs to the third classification level (L3-Temp. classifiers). 
The classifiers that take temporal relations in consideration are performing the best, on average.
The steps of our approach are, in more detail, as follows:

1) Foreground descriptor: create a foreground descriptor that selects in un unsupervised manner top $k$ features and top $q$ frames.
The top $k$ features are selected as the strongest ones, based on their response for each frame. Values are averaged
over the whole video and then the final ordering is done and the top $k$ are selected from
the entire video. Then, using the average value of the selected top-$k$ features as a measure of frame saliency, we
then select the top-$q$ frames with the top $q$ averages of the top-$k$ selected features. 
We consider this top-$q$ group of frames as the most salient
frames and our results confirm the usefulness of the idea. In the experiments section we discuss 
in more detail the effectiveness of the frames and feature selection procedure.
By taking the average responses of the top $k$ features over the top $q$ frames, 
we construct a foreground descriptor over the corresponding video region 
(over which the selection and averaging was performed). This foreground descriptor produces 
the first L1-FG
classifiers in combination with linear SVM, for recognizing subjects, verbs and
objects, individually. 
As mentioned before, note that we do not use any explicit motion features. All our features ( $1000$ in number) 
are collected from the last fully connected layer (fc7) of deep CNNs trained on static images, as discussed before.

2) Foreground-Background descriptor: we augment the foreground descriptor with a descriptor computed from averages over the entire
sequences of frames, minus the selected top-$q$. We show that by augmenting the initial descriptor with the background information the
performance increases significantly. Again we use linear SVMs to obtain our augmented L2-FG-BG classifiers. While feature selection and background
removal helps at the first level, adding background information back and treating it separately, further improves performance at the second level.

3) So far we have not considered temporal ordering, even though we could expect that different parts of the video will be represented
by different classifier outputs. For example, when a child plays with the ball, we could expect the child to appear first, 
then the ball and then
all three classifiers could fire: \emph{child, play and ball} somewhere close to the middle of the video. 
By considering contextual information at different
moments in time we could hope to improve classification even more.
The temporal ordering is considered by computing separate responses for each L2 classifier and for each subject, verb and object over different temporal regions in the video. In these experiments we divide videos in three parts, such that the first part overlaps over half the frames with the middle part, which in turn overlaps for half the frames with the third part.
At this third level we consider top-$c$ responses for subjects, verbs and objects of the L2 classifiers, for the different parts of the video, to obtain a descriptor with $9c$ non-zero values (each of the three video parts will get $c$ responses from the top $c$ subjects, verbs and objects). These will be passed to the final L3-Temp. classifiers. When the temporal information is not considered, then the descriptors will have only $3c$ 
non-zero values (responses are considered over the whole undivided video), to become inputs to the L3-FG-BG classifiers. All classifiers in our system are learned with linear SVMs over the corresponding input features.

\section{Experimental Analysis}

We test our method on the task of Subject-Verb-Object prediction from video
and compare to current methods, which use both visual as well as language
models. We test on the dataset containing YouTube videos made available by 
Chen and Dolan, 2011. This dataset contains $1970$  video clips all having several text
descriptions. We use the same training and testing split as in previous works:
there are $1299$ training videos and $671$ testing videos, as in (Guadarrama et al. 2013).

We evaluate the methods in two ways. First, we test their capability to recognize the exact
subject, verb and object given by the human annotators (Table \ref{tab:most_and_any}). One variant
is to predict the most common word in any extracted human annotated 
triplet. The other variant (shown in parenthesis in
the tables) is to predict any of the words given.
Another way to test the accuracy of the predictions is to evaluate the words given at a higher, more semantic level
of abstraction, by comparing the meanings of the words generated with the ones provided by the human annotators,
by using WUP scores.
Again, we also have two sub-variants, by predicting the most common or any given word, respectively (Table \ref{tab:most_and_any}).

\begin{table}
\caption{Average binary accuracy of predicting the most common word.
The accuracy of the models when their prediction for each sentence component is considered correct only if it is the word (subject, verb and object)
most commonly used by human annotators to describe the video. 
In parenthesis: average binary accuracy of predicting any given word. The accuracy of the models when the prediction is considered correct if used by any of the annotators to describe the video. Note how each level in our contextual, visual story 
hierarchy gives an extra boost in performance.
}
\label{tab:most_and_any}
\begin{center}
\begin{tabular}{lccc}
\toprule
Method          &  S\%    & V\%     & O\%      \\
\midrule
n-gram          & 76.57(86.87) & 11.04(19.25) & 11.19(21.94)  \\
HVC             & 76.57(86.57) & 22.24(38.66) & 11.94(22.09)  \\
FGM             & 76.42(86.27) & 21.34(37.16) & 12.39(24.63)  \\
LSTM-YT         & 71.19(79.40) & 19.40(35.52) & 9.70(20.59)   \\
Guardamma       & \textbf{78.51} & \textbf{22.09}      & \textbf{12.84}       \\
\midrule
L1 - FG         & 74.96(86.18) & 16.98(38.59) & 9.19(25.20)   \\
L2 - FG-BG      & 75.85(86.73) & 19.82(41.87) & 10.58(26.08)  \\
L3 - FG-BG      & 76.61(\textbf{87.12}) & 20.87(41.88) & 12.11(\textbf{26.41})  \\
L3 - Temp.      & 74.51(85.69) & 21.45(\textbf{42.13}) & 11.77(23.98)  \\
\bottomrule
\end{tabular}
\end{center}
\end{table}

\begin{table}
\caption{Average WUP score of predicting most common word - the accuracy of the models when the prediction is considered correct if the word most commonly used by human annotators to describe the video.  In brackets:
average WUP score of predicting any given word - the accuracy of the models when the prediction is considered correct if used by any of the annotators to describe the video.
Note that we outperform the competing methods by a very large margin of about $20\%-30\%$ on verbs and objects. 
We believe the large difference in the WUP case is due to a combination of factors, 
such as efficient foreground frames and feature selection as well as the use of contextual relationships between subjects, verbs, objects and the overall scene. \textbf{Note:} the WUP case is the one in which an answer is correct if it is close, in meaning to the human ground truth, even if it is not the exact same word. Also note how each level in our contextual hierarchy gives an extra boost in performance.
}
\label{tab:most_and_any_WUP}
\begin{center}
\begin{tabular}{lccc}
\toprule
Method          &  S\%    & V\%     & O\%      \\
\midrule
n-gram          & 89.00(96.60) & 41.56(55.08) & 44.01(65.52)  \\
HVC             & 89.09(96.54) & 48.85(65.61) & 43.99(65.32)  \\
FGM             & 89.01(96.32) & 47.05(63.49) & 45.29(67.52)  \\
Guardamma       & 88.94(93.28) & 43.56(59.92) & 36.77(51.91)  \\
\midrule
L1 - FG         & 92.10(\textbf{96.87}) & 69.44(87.48) & 61.77(78.19)   \\
L2 - FG-BG      & 92.39(96.72) & 71.38(88.67) & 61.84(81.22)   \\
L3 - FG-BG      & \textbf{92.51}(96.77) & 71.51(88.71) & 62.09(81.34)   \\
L3 - Temp.      & 92.43(96.74) & \textbf{72.67}(\textbf{89.90}) & \textbf{62.24}(\textbf{81.37})   \\
\bottomrule
\end{tabular}
\end{center}
\end{table}

The tests showed a few interesting properties of our approach versus the others.
First, we see that in the case of raw predictions, all methods perform very similarly
for subjects and objects, while the performance differs more on verbs.
The cause of this is worth
investigating further. We believe it has to do with the fact that the poorly performing methods
do not use movement information, when verbs are guessed from static information alone, as also 
it is the case in our system.

However, when semantic similarity relationships are used at higher levels of interpretation using WUP scores
(Table \ref{tab:most_and_any_WUP}),
we see that most methods
in fact predict words with meanings that are much close to the ground truth, with the same
similar performance for both subjects, object and verbs. Note that the increase in performance
after semantic similarity is considered is very significant. In this case, however, our method
outperforms most of the others by a great margin. This means that our method finds verbs with
much better meanings, even if it does not match the exact word. The accuracy at the level of meaning
comes from the contextual interplay of high level visual features, considered by our approach.
Our experiments validate the intuition that the interactions at higher level of context, occurrence among
strong responding classifiers and temporal ordering is important. It strongly suggests that interactions and spatiotemporal
context should be considered for a fuller understanding of the events taking place in videos.

\paragraph{Foreground feature selection:} selecting the top $k$ features proved to be very effective in our experiments. As we mentioned before, we consider the average feature responses over the whole video and select the features with the top-$k$ average response as the foreground ones. These features represent the output of the $fc7$ layer (of $1000$ values) from the AlexNet-type CNN pre-trained on ImageNet. These features, being at the top of the deep network hierarchy, are highly semantic and are indicative of video content. Thus, 
we expect the features with strong responses to be related to the classes of interest. Instead of using $1000$ features, selecting a small group proved to be much more powerful. Our experiments showed that by selecting the top $60$ features from the video, and zero-ing out the outputs of all the others, the recognition performance increased by a very significant $20\%$ (relative to the recognition rate when all $1000$ were used). Then, by using the average outputs of the selected features, we selected the frames with top-$q$ averages of the top-$k$ features selected. Then, the final foreground video descriptor was computed by averaging (mean pooling) the output of the top-$k$ features only over the selected top-$q$ frames. Values corresponding to all other features in the descriptor were set to $0$. In our experiments, the optimal $q$ validated over the training set was $50$, giving an improvement of extra $10\%$ at test time, over the case when all frames were considered. Therefore, overall, the selection of the most salient frames and features increased our performance by $30\%$ (in relative terms).

\section{Discussion and Conclusions}
We have presented an efficient method for video to language translation, in the form
of SVO triplets, that takes in consideration only visual information by integrating higher level
contextual interactions between foreground  and background, co-occurrence of semantic classifiers as well
as temporal ordering of subjects, objects and verbs. One main novelty of our method is to show that
visual information alone could outperform at a semantic level, more complex models that integrate both vision and
semantics. Another important contribution of our work is to show how contextual relationships, at different levels of visual interpretation, can boost recognition performance, which automatically produces more coherent translations. We concentrate on the general idea of a visual story, in which objects, scenes and interactions are understood in a unified manner. We lay down its basic principles and present an efficient and relatively simple implementation that significantly outperforms more complex systems, such as those based on Long-Short-Term-Memory neural networks.
Our paper, by focusing on both visual aspects of interpretation and contextual relationships, could open important doors towards a full, coherent understanding of the visual world, which will also facilitate the translation into natural language.

\section{Acknowledgments}
This work was supported in part by CNCS-UEFICSDI, under project PNII PCE-2012-4-0581.
The authors would like to thank Dr.~Rahul Sukthankar for numerous fruitful discussions and
helpful feedback.

\bibliography{iclr2016_conference_story}
\bibliographystyle{iclr2016_conference}

\end{document}